\documentclass{article}

\usepackage[preprint,nonatbib]{nips_2018}

\usepackage[utf8]{inputenc} 
\usepackage[T1]{fontenc}    
\usepackage{hyperref}       
\usepackage{url}            
\usepackage{booktabs}       
\usepackage{amsfonts}       
\usepackage{nicefrac}       
\usepackage{microtype}      
\usepackage{graphicx}

\usepackage{comment}
\newcommand{\JP}[1]{{\color{red}{[}\textbf{Jeff says: #1}{]}}}

\title{Structurally Sparsified Backward Propagation for Faster Long Short-Term Memory Training}

\author{
  Maohua Zhu, Yuan Xie \\
  Department of Electrical and Computer Engineering\\
  University of California, Santa Barbara\\
  \texttt{\{maohuazhu,yuanxie\}@ece.ucsb.edu} \\
  \AND
  Jason Clemons, Jeff Pool, Stephen W. Keckler \\
  NVIDIA \\
  \texttt{\{jclemons,jpool,skeckler\}@nvidia.com} \\
  \AND
  Minsoo Rhu \\
  Department of Computer Science and Engineering \\
  POSTECH\\
  \texttt{minsoo.rhu@gmail.com} \\
}

\begin{document}
\maketitle

\begin{abstract}
Exploiting sparsity enables hardware systems to run neural networks faster and more energy-efficiently. However, most prior sparsity-centric optimization techniques only accelerate the forward pass of neural networks and usually require an even longer training process with iterative pruning and retraining. We observe that artificially inducing sparsity in the gradients of the gates in an LSTM cell has little impact on the training quality. Further, we can enforce structured sparsity in the gate gradients to make the LSTM backward pass up to 45\% faster than the state-of-the-art dense approach and 168\% faster than the state-of-the-art sparsifying method on modern GPUs. Though the structured sparsifying method can impact the accuracy of a model, this performance gap can be eliminated by mixing our sparse training method and the standard dense training method. Experimental results show that the mixed method can achieve comparable results in a shorter time span than using purely dense training. 
\end{abstract}

\vspace{-15pt}
\section{Introduction}
Long Short-Term Memories (LSTMs)~\cite{Hochreiter1997} are an important type of Recurrent Neural Network (RNN) that are widely used to process sequential data. While the complex neuron structure of an LSTM mitigates exploding or vanishing gradients, the correspondingly large amount of computation makes the training process time-consuming.

Exploiting sparsity in the processing of neural networks has led to improved computational and energy performance with minimal impact on the accuracy of results. Common sparsity-centric optimization techniques include pruning insignificant weight parameters~\cite{HanPruning2015,wen2016learning,han2017ese} and dynamically skipping zero activations~\cite{Albericio2016}. The weight pruning approach can dramatically reduce the size of network models and achieve significant speedup over the original fully dense network implementation. However, these optimizations are primarily designed to be used during forward propagation, and such techniques often require complicated and time-consuming iterative retraining which increases the overall time to train a neural network. The zero skipping approach elides the multiplication and addition operations that would take place when activation operand is zero. This occurs quite often in feed-forward networks that use  the ReLU activation function in the forward propagation such as Convolutional Neural Networks (CNNs). However, LSTM cells typically do not use ReLU and thus often have many non-zero values in their activations. 

Instead of exploiting the sparsity in the forward propagation, recent works show that aggressively quantizing or sparsifying the gradients of deep neural networks can minimize the communication overhead in a distributed training system~\cite{seide20141,wen2017terngrad,lin2017deep}. Sparsifying the gradients of LSTM can also improve the training speed with little impact on the training quality due to its effect of regularization~\cite{sun2017meprop}. However, such sparsifying method requires lower enforced sparsity level for the quality of training large LSTM models. Unfortunately, existing works fail to exploit moderate sparsity, such as 50\%, to achieve better LSTM training performance than the dense counterpart on modern compute engines such as GPUs, for high sparsifying and sparse matrix multiplication overhead. 

To tackle these problems in existing gradient sparsifying, we explore hardware-oriented structural sparsifying methods that allow us to exploit the fast hardware primitives on GPUs, leading to an maximum 45\% faster LSTM backward propagation in practice. While accuracy loss can be regained with extra training, this could defeat the purpose of accelerating training. Instead, inspired by past work~\cite{han2016dsd}, we show that by using this sparse phase for \textit{most} of the training with a brief unmodified (dense) phase to finish the original training schedule, networks can spend less time in training.

\section{Background and Motivation}
In this section, we first review some of the prior work on sparsity-centric optimization techniques for neural networks, and then illustrate the observation of potential sparsity in the LSTM backward propagation.

\subsection{Sparsity Optimization for Neural Networks}
~\cite{Denil2013} demonstrated that there is significant redundancy in the parameterization of deep neural networks. Consequently, the over-sized parameter space results in sparsity in the weight parameters of a neural network. Besides the parameters, there is also sparsity in the activations of each layer in a network, which comes from the activation function of neurons, such as ReLU~\cite{Krizhevsky2012}. 

As the sparsity in weight parameters do not depend on the input data, it is often referred to as \textit{static sparsity}. On the other hand, the sparsity in the activations depend on not only the weight values but also the input data. Therefore, we refer to the sparsity in the activations as \textit{dynamic sparsity}.

Exploiting sparsity can dramatically reduce the network size and thus improve the computing performance and energy efficiency. ~\cite{HanPruning2015} applied network pruning to Convolutional Neural Networks (CNNs) to significantly reduce the footprint of the weights, which can enable all the weights to be stored in SRAM. 
However, weight sparsity is commonly used to accelerate only inference or the forward pass of training. Fortunately, leveraging the dynamic sparsity can benefit both the forward and backward passes of the training and inference phases of neural networks. Recent publications have proposed various approaches to eliminate ineffectual MAC operations with zero operands~\cite{han2016eie,han2017ese,Albericio2016}. Although these sparsity-centric optimization approaches have achieved promising results on CNNs, much less attention has been paid to LSTM-based RNNs, because there is a common belief that the major source of the dynamic sparsity is the ReLU function, which is widely used in CNNs but not in LSTMs. To accelerate LSTM RNNs, we want to exploit any opportunities to use dynamic sparsity in the LSTM training process.

\subsection{Quantizing and Sparsifying the Gradients}\label{ssec:sparsify-gradient}
It is not new that the redundancy in the gradients of the neural networks can be exploited to minimize the communication overhead in distributed deep learning systems. \cite{seide20141} aggressively quantized the gradients to 1 bit per value to shrink the data size to be synchronized. TernGrad~\cite{wen2017terngrad} uses only three numerical levels to represent gradients and theoretically proves the convergence under this constraint. These aggressive quantization does not incur accuracy loss on AlexNet~\cite{Krizhevsky2012} and induces only 2\% on average for GoogLeNet~\cite{szegedy2015going}. 

Unfortunately, quantizing gradients helps little with accelerating the training phase since modern compute engines only support standard precisions such as FP32, FP16, and INT8. To reduce computations, meProp~\cite{sun2017meprop} proposes to sparsify the gradients. On one hand, the meProp sparsifying approach significantly reduces the total number of MAC operations required for training. On the other hand, the sparsifying acts as a regularizer in the backpropagation so that the training still converges for small datasets such as MNIST. 

However, meProp is only feasible for accelerating LSTM models that are trained with small datasets. When datasets become large, high sparsity approaches, such as keeping only top 10\% gradients, suffer catastrophic accuracy loss. To avoid the accuracy loss, we have to lower the sparsity enforced on the gradients. But such moderate sparsity, e.g. 50\%, makes meProp slower than the corresponding dense matrix multiplications on the latest GPUs (see Figure \ref{fig:speedup}). This situation is caused by two reasons: 1) the global top $k$ operation becomes very time-consuming when $k$ is large or sparsity is low, and 2) meProp cannot exploit the hardware primitives for matrix multiplications introduced on the latest GPU architecture. To tackle these problems, we propose a novel structural sparsifying methods to accelerate large NN workloads.

\section{Structural Sparsifying the Gate Gradients}

\begin{figure}[t]
\begin{center}
\includegraphics[width=0.6\textwidth]{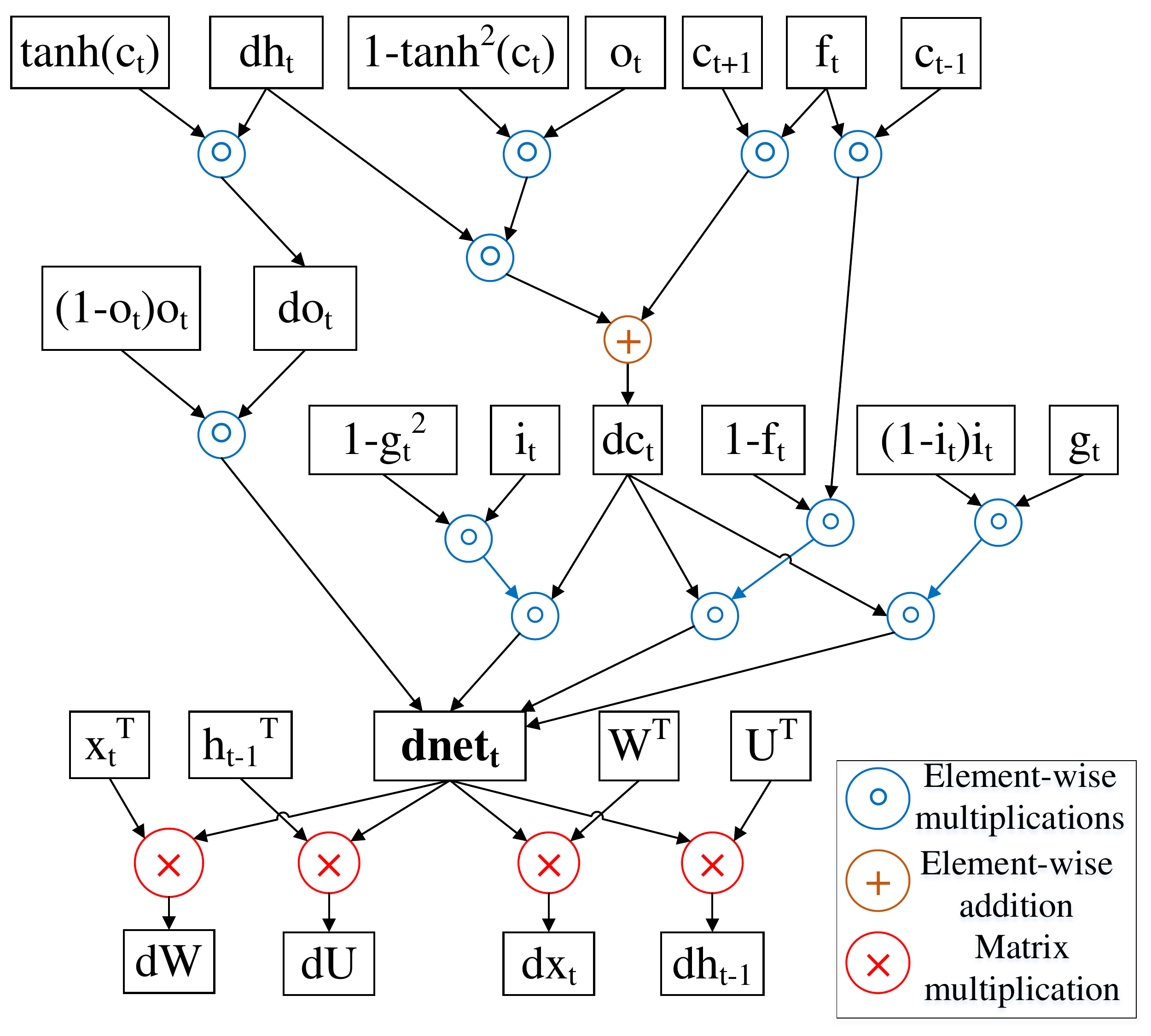}
\end{center}
\vspace{-15pt}
\caption{LSTM backward pass data flow. The sign "$\circ$" signifies element-wise vector multiplication. "$\times$" stands for matrix multiplication. Biases are omitted for simplicity.}\label{fig:LSTM-BP}
\end{figure}

As discussed in Section \ref{ssec:sparsify-gradient}, traditional gradient sparsifying approaches, such as meProp, suffer accuracy loss if applied to large-scale LSTM models. Figure \ref{fig:LSTM-BP} shows the data flow of the LSTM backward propagation. Traditional gradient sparsifying methods remove unimportant values in output gradient ($dh_t$ in Figure \ref{fig:LSTM-BP}), which is a sub-optimal target. We observe that the gate gradient ($dnet_t$ in Figure \ref{fig:LSTM-BP}) is more resilient to sparsifying than $dh_t$ and directly used by the most time-consuming matrix multiplication part. Specifically, the $dnet$ is involved in all the four matrix multiplications so that we can greatly reduce the amount of computation by sparsifying $dnet$. Therefore, we choose $dnet$ as our sparsifying target instead of $dh$.  



meProp~\cite{sun2017meprop} removes entire rows or columns so they have a smaller, dense computation to perform. However, removing the rows or columns to form smaller matrices is costly on GPUs and offsets the speedup when the sparsity is not extremely high. Furthermore, the global top k computation used in meProp is costly when the sparsity is relatively low. To get rid of row/column regrouping and global top k computations, we propose a hardware-oriented structural sparsifying method for the LSTM backward propagation. The structural sparsifying method enforces a fixed level of sparsity in the LSTM gate gradients, yielding sparse gradient matrices useful for block based matrix multiplications. 

\begin{figure}[ht]
\begin{center}
\includegraphics[width=0.95\textwidth]{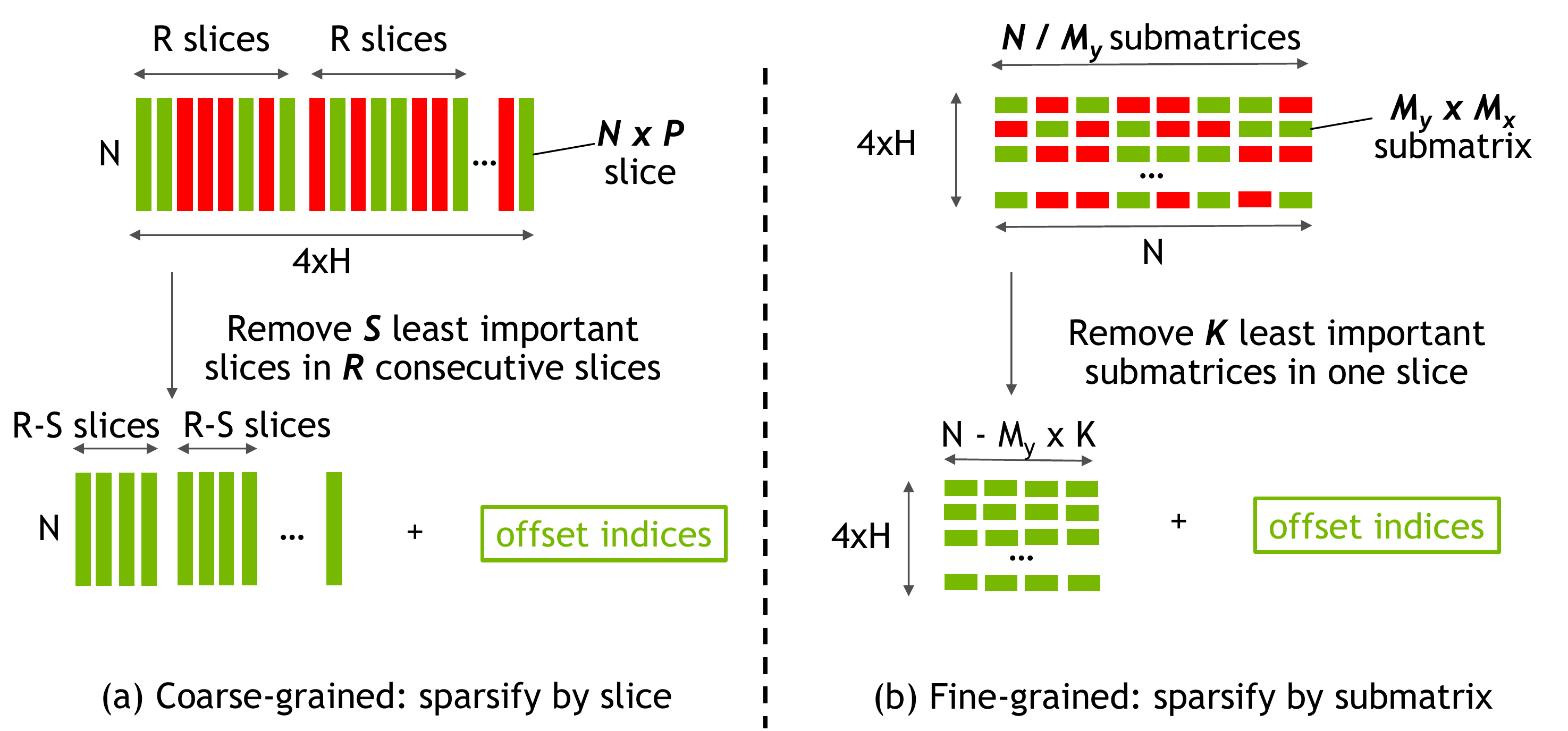}
\end{center}
\caption{Examples of structural sparsifying: (a) coarse-grain sparsifying and (b) fine-grain sparsifying. Matrices in (b) are transposed for better view. }\label{fig:sparsifytensorcore}
\end{figure}

Figure \ref{fig:sparsifytensorcore} shows two versions of the structural sparsifying methods, coarse-grained sparsifying and fine-grained sparsifying, for an layer of $H$ LSTM cells being trained with a mini-batch size of $N$. Instead of thresholding based on the absolute value of an element in the gate gradient matrix, the structural sparsifying methods evaluate a larger regional block. 

The coarse-grained structural sparsifying method first splits the gate gradient matrix evenly into multiple slices and then evaluates the L2 norm of each slice. Instead of splitting a matrix into single columns or rows, our slice consists of $P$ consecutive columns (or rows) to satisfy the hardware constraint to exploit matrix multiplication primitives without the costly row/column regrouping. Then the adjacent slices are sequentially grouped into {\it sparsifying region}. In the example shown on the left in Figure \ref{fig:sparsifytensorcore}, a $N\times 4 \times H$ gate gradient matrix is divided into $H$ $P$-element-wide slices. Then every $R$ slices are grouped together as a {\it sparsifying region}. Based on the evaluated L2 norm of the slices within a sparsifying region, a fixed level of sparsity can be enforced by removing the least important slices. For example, 50\% sparsity is enforced when $S=R/2$. 


After the least important slices being removed, the sparsified gate gradient matrix can be represented in a dense matrix format with an additional {\it offset index array} to record the slices that are kept. This feature enables the sparsified matrix to be efficiently computed on GPUs using block based matrix multiplication. 

However, the strict constraint of the coarse-grained sparsifying potentially has impact on the training quality since the sparsifying region contains at least a column or a row of the gate gradient matrix. To increase the flexibility of the structural sparsifying, we propose a fine-grained sparsifying scheme, which is shown on the right in Figure \ref{fig:sparsifytensorcore}. The fine-grained sparsifying further breaks the gate gradient matrix from a slice into multiple $M_y\times M_x$ submatrices. This is because, as mentioned above, the block based matrix multiplication uses fast matrix multiplication primitives which consume a $M_y\times M_x$ operand matrix. Meanwhile, the sparsifying region size is reduced to a slice of $N\times M_x$ to keep the index array to an acceptable size. The fine-grained sparsifying increases the flexibility but breaks the high performance coalescing memory fetch pattern, which potentially degrades the performance of the following matrix multiplication.

\section{Experimental Methodology}\label{sec:methodology}

Our structural sparsifying methods are designed to accelerate the LSTM backward pass with minimal effect on the quality of training results. To evaluate both the training quality and speedup over the typical dense backward pass of LSTM training, we trained three LSTM based models with different configurations of our structural sparsifying methods on a GPU accelerated server and compared the results to using the typical dense method on the same machine. 
The server has an NVIDIA Tesla V100 GPU~\cite{volta} which is based on the Volta architecture. The Volta architecture includes the new TensorCores, which provide  hardware support for fast block based matrix multiplication primitives. Matrix multiplications of specific sizes can be executed on specialized hardware, which enables the structural sparsified training achieves higher performance. Based on the functionality of TensorCore, we set $P=4$ for the coarse-grained method and $M_y=8$, $M_x=4$ for the fine-grained method to utilize the fast matrix multiplication primitives effectively. 

To ensure the generality of the sparsification techniques,  we evaluated them using applications in three domains: language modeling, machine translation, and image captioning.
The word language model has two LSTM layers as described in ~\cite{zaremba2014recurrent}. The dataset used for the word language model is Penn Tree Bank (PTB)~\cite{marcus1993building}. Each of the LSTM layers has 512 neurons and the size of each minibatch is 64. We train the LSTM for 40 epochs with a learning rate of 1. All other settings are the same as ~\cite{zaremba2014recurrent} and can be found in the TensorFlow tutorial example~\cite{repo:word-lm}.  

We use the Show and Tell model~\cite{vinyals2017show} for the image captioning experiment. The image captioning model consists of an Inception V3 model~\cite{szegedy2016rethinking} with an LSTM layer attached to the last layer of the CNN\@. The LSTM layer has 512 cells by default. Since our sparsifying methods only apply to the LSTM layer, we use a pre-trained Inception V3 model and randomly initialize the parameters of the LSTM layer. The training dataset is MSCOCO~\cite{Lin2014} and the mini-batch size is set to 64. We trained the model for 500,000 steps (about 55 epochs by default configuration). To evaluate the quality of generated captions, we calculated the BLEU-4 score for each training configuration on the MSCOCO test dataset. The code we used can be found in TensorFlow model zoo~\cite{repo:im2txt}.

For the machine translation application, we trained an encoder-decoder architecture with an attention mechanism to perform Neural Machine Translation (NMT)~\cite{britz2017massive, Luong2015effective}. We use an architecture with a 2-layer LSTM encoder, a 4-layer LSTM decoder, and an attention module. The first layer of the LSTM encoder is bidirectional and the other LSTM layers are unidirectional. Both the unidirectional and bidirectional layers have 512 LSTM cells. In the experiments, we use the BLEU score as the metrics of the neural machine translation model. The dataset for the training task is the WMT 16 English-German dataset. Since we observed that the validation BLEU score changes little after 600,000 steps (about 9 epochs), we only use the first 16 epochs to evaluate the training quality. 
Beyond this simplification, we followed the instructions to reproduce the NMT training with an open source framework~\cite{repo:nmt}.



The neural network models are implemented with TensorFlow v1.2, which uses the CUBLAS library~\cite{cublas} for matrix multiplications.  We use CUBLAS as the baseline for our experiments but then added a custom sparsifying operation to TensorFlow to implement structured sparsity. The forward pass of the custom operation uses CUBLAS matrix multiplication to compute the activations of the gates. The backward pass of the custom operation first structurally sparsifies the gate gradients, and then computes the weight gradients and data gradients using the sparsified gate gradients. Because CUBLAS is a closed library and could not be used for the sparsified matrix multiplication, we leveraged the open source CUTLASS library~\cite{cutlass} for the sparsified matrix multiplication operations and augmented it with the required indexing code to support the structurally sparsified gate gradient matrices. In our implementation, the offset index array is integrated in the associated sparsified matrix, storing the index in the least significant bits of the mantissa of the floating point numbers. Our experiments require at most 5 bits to store each index; our results show that this arrangement has negligible effect on the precision of the 32-bit floating-point numbers. To compare our methods with meProp~\cite{sun2017meprop}, we implemented their unified top k method, which is used for their speedup evaluation, with different sparsity levels. Other configurations for meProp are the same as our sparsifying methods.

\section{Experimental Results}

We present the LSTM training results with and without our structural sparsifying method. Although the performance gap between the sparse training and the dense training is small, the quality decrease can be mitigated by augmenting the sparse training steps with the dense training steps. We show that doing this can lead to shorter training time for our target LSTM based networks.

\subsection{Language Modeling}\label{ssec:dse}

Table \ref{tab:lm-result} shows the test perplexity of the language models trained with various configurations of the coarse-grained method, the fine-grained method, and the regular dense method. The sparsity enforced by the coarse-grained method can be controlled by the parameters $R$ and $S$\@. For example, if we set $R=8$ and $S=4$, the sparsity enforced by the coarse-grained method is 50\% since half number of the slices within a sparsifying region are removed. For the fine-grained method the sparsity is controlled by setting $K$\@. For example, we can enforce 50\% sparsity by setting $K=4$ since each slice contains $8$ submatrices given the problem size of $N=64$ and $H=512$.

\begin{table}
\centering
\caption{Performance of trained language models using different configurations. Lower perplexity means better performance.}
\label{tab:lm-result}
\begin{small}
\renewcommand\tabcolsep{3pt}
\begin{tabular}{llccc}
\toprule
 Training Method & Configuration & Sparsity & Perplexity & Improvement \\
\midrule
Dense      & N/A & 0\% & 93.101            & 0                              \\ \hline
Coarse & R=4, S=2 & 50\% & 99.43            & -6.80\%  \\ \hline
Coarse & R=8, S=2 & 25\% & 94.39            & -1.39\%                         \\ \hline
Coarse & R=8, S=4 & 50\% & 98.61            & -5.92\%                         \\ \hline
Coarse & R=8, S=6 & 75\% & 112.84            & -21.20\%                         \\ \hline
Coarse & R=16, S=8 & 50\% & 98.58            & -5.88\%  \\ \hline
Coarse & R=32, S=16 & 50\% & 98.11            & -5.37\%  \\ \hline
Fine & K=2 & 25\% & 93.079            & +0.02\%                         \\ \hline
Fine & K=4 & 50\% & 96.410            & -6.48\%                         \\ \hline
Fine & K=6 & 75\% & 108.046            & -16.05\%                         \\ 
\bottomrule
\end{tabular}
\end{small}
\end{table}

Table \ref{tab:lm-result} shows that the resulted models have comparable performance with the baseline dense model when only 25\% sparsity is enforced for both coarse-grained and fine-grained method. If the sparsity is increased to 50\%, the performance difference is increased to 6\%. However, if the sparsity is further increased to 75\%, the resulted models will be unacceptably degraded to more than 20\%.
The fine-grained method leads to a model with higher performance than the coarse-grain method because the fine-grain method is more flexible. Since it works on smaller regions of the gate gradients, it can better avoid removing significant individual gradient elements.  


As a compromise between the quality and speed of the training, we choose 50\% sparsity for further study. For the coarse-grained method, the sparsifying region can be adjusted without large consideration of the underlying matrix computation mechanisms. However, the optimal sparsifying region of the fine-grained method is determined by the problem size and the hardware/software platform characteristics. Table \ref{tab:lm-result} shows the results of model performance with different sparsifying region sizes. Every scheme enforces 50\% sparsity but larger sparsifying regions results in slightly better model in general. However, the benefit from increasing the sparsifying region diminishes when $R>8$. Therefore, we choose $R=8, S=4$ for further experiments to optimize for less computation and accuracy loss.

\subsection{Dense After Sparse Training}

Although the models trained with the sparse methods suffer only a small decrease in accuracy, some quality-sensitive applications still require a comparable accuracy with the dense method results. DSD~\cite{han2016dsd} demonstrated combining pruning and dense retraining can lead to better accuracy than pure dense training. Inspired by this work, we propose a dense after sparse training method, which combines our sparse training method with the regular dense method to compensate for degradations in quality.

\begin{table}
\centering
\caption{Performance of trained language models by dense after sparse method. Lower perplexity means better performance.}
\label{tab:sparse-dense}
\begin{small}
\begin{tabular}{lcc}
\toprule
Training Method & Perplexity & Improvement \\ \midrule
Dense      & 93.101            & 0                              \\ \hline
Coarse-grained & 98.611     & -5.92\%                         \\ \hline
1/2 coarse + 1/2 dense & 88.675 & +4.75\%  \\ \hline
3/4 coarse + 1/4 dense & 91.388            & +1.84\%                         \\ \hline
5/6 coarse + 1/6 dense & 99.47             & -6.84\%                         \\ \hline
Fine-grained & 96.410            & -3.55\%                         \\ \hline
1/2 fine + 1/2 dense & 88.607 & +4.83\%                          \\ \hline
3/4 fine + 1/4 dense & 91.118            & +2.13\%                         \\ \hline
5/6 fine + 1/6 dense & 96.151            & -3.28\%                         \\ 
\bottomrule
\end{tabular}
\end{small}
\end{table}

Table \ref{tab:sparse-dense} shows the results of the dense after sparse training. The model is first trained with the coarse-grained method or the fine-grained method, and then trained with the regular dense training model. With 75\% steps of sparse training and 25\% steps of dense training, the resulted model achieves slightly better performance than the baseline model.  The number of total training steps remains constant across the methods. These results demonstrate that mixing the two methods compensates for  the quality gap between the pure sparse training and pure dense training methods. A key parameter of applying this mixed training is the ratio of sparse steps to dense steps. Intuitively, the resulting model will perform better with more dense time. Our results show that a 3-to-1 ratio of sparse to dense is sufficient. 

To evaluate the generality of our sparsifying methods, we apply both our coarse and fine grained sparsifying techniques to our other benchmarks. We use settings to enforce 50\% sparsity while varying the ratio of dense to sparse training.

\subsection{Image Captioning}

As we only train the LSTM of the Show and Tell model, we chose to cache the outputs of the Inception V3 to avoid the need to perform the inference pass on the CNN\@. The configuration of the coarse-grained method is $R=8, S=4$. The configuration of the fine-grained method is $K=4$. Both configurations enforce 50\% sparsity in the gate gradients. 
%
\begin{table}
\centering
\caption{Performance of trained image captioning models. Higher BLEU-4 score means better performance.}
\label{tab:image-captioning}
\begin{small}
\begin{tabular}{lcc}
\toprule
Training Method & BLEU-4 & Improvement \\ \midrule
Dense      & 31.0            & 0                              \\ \hline
Coarse-grained & 30.3     & -2.26\%                         \\ \hline
3/4 coarse + 1/4 dense & 30.8            & -0.65\%                         \\ \hline
Fine-grained & 30.6            & -1.29\%                         \\ \hline
3/4 fine + 1/4 dense & 31.1            & +0.32\%                         \\ \bottomrule
\end{tabular}
\end{small}
\vspace{-5pt}
\end{table}

Table \ref{tab:image-captioning} shows that both coarse-grained and fine-grained methods degrade quality by less than 3\%. Compared to the 5.92\% performance loss in language modeling, this result demonstrates generality to applications beyond language model datasets. Moreover, the performance gap between the sparse training and the dense training can be mitigated by the dense after sparse approach. Interestingly, the 3-to-1 sparse-to-dense ratio used in the word language model also works well for the image captioning.

\subsection{Machine Translation}

Table \ref{tab:nmt-bleu} shows the performance on the validation datasets of the models trained with both dense SGD and our structurally sparsified training methods. The BLEU scores are the validation BLEU scores at the end of the training (after 600,000 steps). Although our sparsifying methods enforce 50\% sparsity in the gate gradients, the resulting models achieve acceptable BLEU scores. Even compared to the dense SGD, the coarse-grain sparsifying only suffers a 3.5\% BLEU score decrease while the fine-grain method is slightly better with just a 2.0\% decrease.

\begin{table}
\centering
\caption{Performance of trained NMT models. Higher BLEU score means better performance.}
\label{tab:nmt-bleu}
\begin{small}
\begin{tabular}{lcc}
\toprule
Training Method & BLEU & Improvement \\ \midrule
Dense      & 20.32            & 0                              \\ \hline
Coarse-grained & 19.60     & -3.5\%                         \\ \hline
3/4 coarse + 1/4 dense & 20.30            & -0.1\%                         \\ \hline
5/6 coarse + 1/6 dense & 20.18 	& -0.7\% \\ \hline
Fine-grained & 19.92            & -2.0\%                         \\ \hline
3/4 fine + 1/4 dense & 20.45            & +0.64\%                         \\ \hline
5/6 fine + 1/6 dense & 20.17 	& -0.74\% \\ \bottomrule
\end{tabular}
\end{small}

\end{table}

The dense after sparse approach works well for the machine translation task. With 50\% sparsity and 75\% sparse training time, the achieved BLEU scores from both the coarse-grained and fine-grained methods are almost the same as the baseline. If the translation quality is not sensitive, we can further increase the sparse time to obtain higher overall speedup. For example, the BLEU score is only 0.7\% lower than the baseline when the coarse-grained sparse training steps are increase to 5/6 of the total steps.

\subsection{GPU Training System Performance Analysis}
\begin{figure}[ht]
\begin{center}
\includegraphics[width=0.9\textwidth]{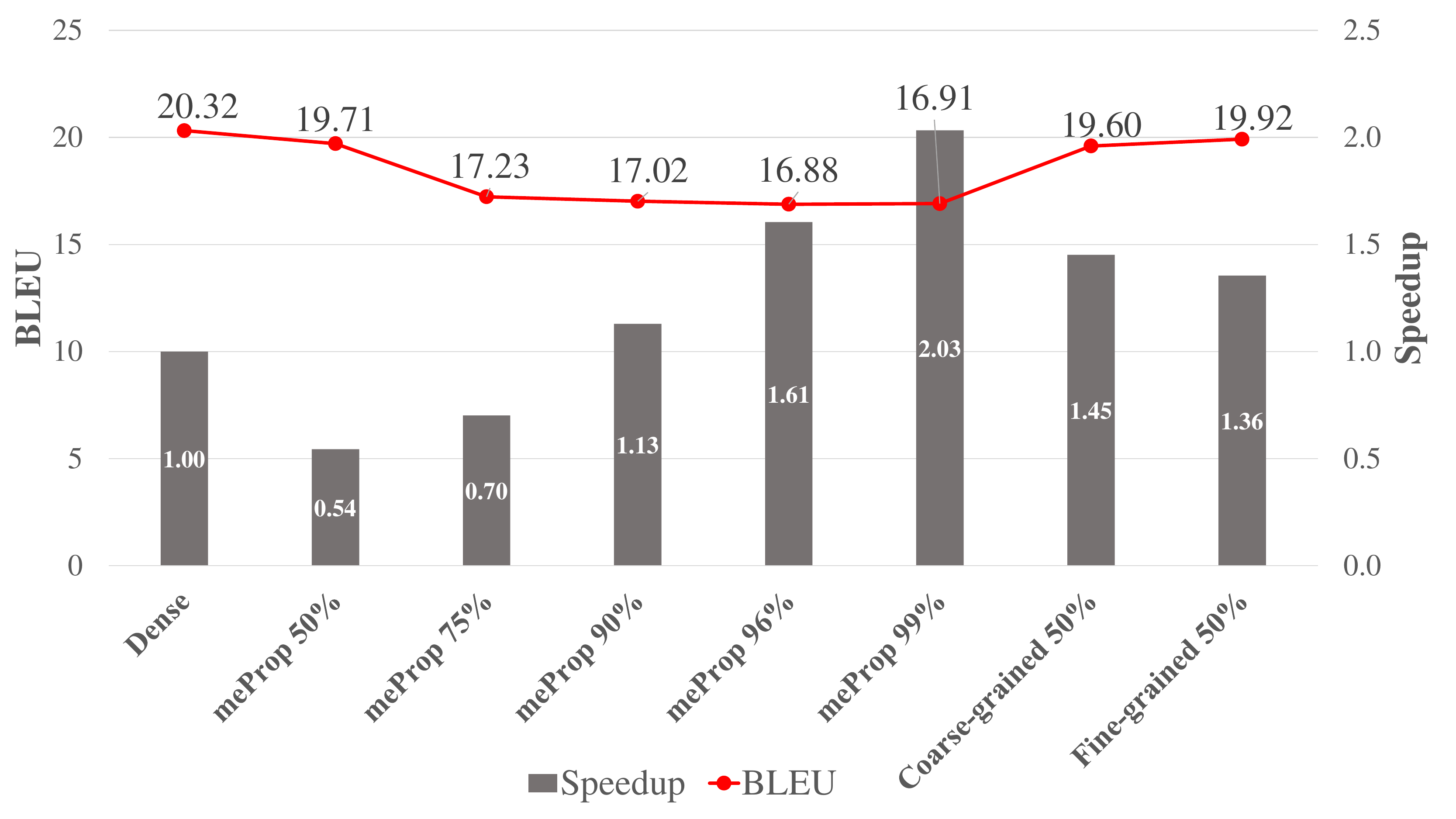}
\end{center}

\caption{BLEU scores of trained models with sparsifying methods and corresponding speedup for LSTM training in NMT. The dense results are based on CUBLAS. We implement meProp with different sparsity levels.}\label{fig:speedup}
\end{figure}

Figure \ref{fig:speedup} shows that the LSTM backward propagation trained with the coarse-grain sparsifying method is 45\% faster than the regular training method while the fine-grain sparsifying is 36\% faster. If we keep top 50\% values, meProp has slightly better quality than our coarse-grained method but worse than our fine-grained method. However, meProp is slower than the dense training if keeping more than 25\% values. Especially, for the 50\% sparsity case which achieves comparable accuracy with respect to the baseline, our coarse-grained method is 168\% faster than the meProp approach. This is because meProp cannot leverage the high-performance matrix multiplication primitives and the unified top k is time-consuming. Note that the speedup numbers only depend on the LSTM network topology and are independent from the application types, which means our sparsifying methods can achieve the same speedups for the LSTM backward pass of language modeling and image captioning. Our results indicate that the high sparsity numbers which enables meProp to achieve good speedup lead to catastrophic performance loss in larger models such as NMT.

\subsection{Discussion}

The experiments show that our sparsifying methods have little impact on the training quality while decreasing the time to train LSTM models when compared to dense training. Because of the structural nature, the speedup depends only on the problem size and the sparsity level, which are both predetermined before training. Thus our sparsifying methods offer the flexibility to trade training quality for training speed. In general, our fine-grained approach allows for more aggressive sparsity settings, but is less efficient, while our coarse technique is faster, but cannot be used as aggressively.

Furthermore, the small performance gap can be mitigated by mixing the dense training with the sparse training, at the cost of less speedup. But since we do not increase the number of total training steps, the average execution time of one step is still shorter than the dense method. For example, with 1/4 dense steps following 3/4 coarse-grained steps, the overall speedup is 1.34x, and the BLEU score is slightly higher than the baseline. Unlike many prior works, our sparsifying methods are tested on large-scale problems and achieve good speedup against the state-of-the-art training hardware and software. Opportunities for more performance exist with more engineering effort spent tuning the sparse kernels.


\section{Conclusion}

In this work, we propose two versions of structural sparsifying methods that enforce a fixed level of sparsity in the gate gradients in the LSTM backward propagation. Training with only the sparsified gate gradients can lead to a minor model performance decrease, we can compensate the performance gap by mixing the sparse training with the regular dense training. With the dense-after-sparse training method, we can train the LSTM models to the same quality as models using the regular dense training with the same number of training steps. As the sparsifying methods reduce the amount of computation, the sparse LSTM backward propagation is 49\% faster than the dense counterpart on GPUs. Experiments also demonstrate the dense after sparse training method can achieve comparable results in a shorter time span than using the pure dense training for large-scale models.  

\bibliographystyle{ieeetr}
\bibliography{reference}

\begin{thebibliography}{10}

\bibitem{Hochreiter1997}
S.~Hochreiter and J.~Schmidhuber, ``Long short-term memory,'' {\em Neural
  computation}, vol.~9, no.~8, pp.~1735--1780, 1997.

\bibitem{HanPruning2015}
S.~Han, J.~Pool, J.~Tran, and W.~Dally, ``Learning both weights and connections
  for efficient neural network,'' in {\em Advances in neural information
  processing systems}, pp.~1135--1143, 2015.

\bibitem{wen2016learning}
W.~Wen, C.~Wu, Y.~Wang, Y.~Chen, and H.~Li, ``Learning structured sparsity in
  deep neural networks,'' in {\em Advances in Neural Information Processing
  Systems}, pp.~2074--2082, 2016.

\bibitem{han2017ese}
S.~Han, J.~Kang, H.~Mao, Y.~Hu, X.~Li, Y.~Li, D.~Xie, H.~Luo, S.~Yao, Y.~Wang,
  {\em et~al.}, ``{ESE}: Efficient speech recognition engine with sparse {LSTM}
  on {FPGA},'' in {\em Proceedings of the 2017 ACM/SIGDA International
  Symposium on Field-Programmable Gate Arrays}, pp.~75--84, ACM, 2017.

\bibitem{Albericio2016}
J.~Albericio, P.~Judd, T.~Hetherington, T.~Aamodt, N.~E. Jerger, and
  A.~Moshovos, ``Cnvlutin: Ineffectual-neuron-free deep neural network
  computing,'' in {\em Computer Architecture (ISCA), 2016 ACM/IEEE 43rd Annual
  International Symposium on}, pp.~1--13, IEEE, 2016.

\bibitem{seide20141}
F.~Seide, H.~Fu, J.~Droppo, G.~Li, and D.~Yu, ``1-bit stochastic gradient
  descent and its application to data-parallel distributed training of speech
  dnns,'' in {\em Fifteenth Annual Conference of the International Speech
  Communication Association}, 2014.

\bibitem{wen2017terngrad}
W.~Wen, C.~Xu, F.~Yan, C.~Wu, Y.~Wang, Y.~Chen, and H.~Li, ``Terngrad: Ternary
  gradients to reduce communication in distributed deep learning,'' in {\em
  Advances in Neural Information Processing Systems}, pp.~1508--1518, 2017.

\bibitem{lin2017deep}
Y.~Lin, S.~Han, H.~Mao, Y.~Wang, and W.~J. Dally, ``Deep gradient compression:
  Reducing the communication bandwidth for distributed training,'' {\em
  International Conference on Learning Representations (ICLR)}, 2018.

\bibitem{sun2017meprop}
X.~Sun, X.~Ren, S.~Ma, and H.~Wang, ``meprop: Sparsified back propagation for
  accelerated deep learning with reduced overfitting,'' in {\em International
  Conference on Machine Learning}, pp.~3299--3308, 2017.

\bibitem{han2016dsd}
S.~Han, J.~Pool, S.~Narang, H.~Mao, E.~Gong, S.~Tang, E.~Elsen, P.~Vajda,
  M.~Paluri, J.~Tran, {\em et~al.}, ``Dsd: Dense-sparse-dense training for deep
  neural networks,'' {\em International Conference on Learning Representations
  (ICLR)}, 2017.

\bibitem{Denil2013}
M.~Denil, B.~Shakibi, L.~Dinh, N.~de~Freitas, {\em et~al.}, ``Predicting
  parameters in deep learning,'' in {\em Advances in Neural Information
  Processing Systems}, pp.~2148--2156, 2013.

\bibitem{Krizhevsky2012}
A.~Krizhevsky, I.~Sutskever, and G.~E. Hinton, ``Imagenet classification with
  deep convolutional neural networks,'' in {\em Advances in neural information
  processing systems}, pp.~1097--1105, 2012.

\bibitem{han2016eie}
S.~Han, X.~Liu, H.~Mao, J.~Pu, A.~Pedram, M.~A. Horowitz, and W.~J. Dally,
  ``{EIE}: {E}fficient inference engine on compressed deep neural network,'' in
  {\em Proceedings of the 43rd International Symposium on Computer
  Architecture}, pp.~243--254, IEEE Press, 2016.

\bibitem{szegedy2015going}
C.~Szegedy, W.~Liu, Y.~Jia, P.~Sermanet, S.~Reed, D.~Anguelov, D.~Erhan,
  V.~Vanhoucke, and A.~Rabinovich, ``Going deeper with convolutions,'' in {\em
  Proceedings of the IEEE conference on computer vision and pattern
  recognition}, pp.~1--9, 2015.

\bibitem{volta}
NVIDIA, ``{NVIDIA} {T}esla {V}100 {GPU} architecture,'' tech. rep., 2017.

\bibitem{zaremba2014recurrent}
W.~Zaremba, I.~Sutskever, and O.~Vinyals, ``Recurrent neural network
  regularization,'' {\em arXiv preprint arXiv:1409.2329}, 2014.

\bibitem{marcus1993building}
M.~P. Marcus, M.~A. Marcinkiewicz, and B.~Santorini, ``Building a large
  annotated corpus of english: The penn treebank,'' {\em Computational
  linguistics}, vol.~19, no.~2, pp.~313--330, 1993.

\bibitem{repo:word-lm}
Google, ``Tensorflow tutorial example: Word language model.''

\bibitem{vinyals2017show}
O.~Vinyals, A.~Toshev, S.~Bengio, and D.~Erhan, ``Show and tell: Lessons
  learned from the 2015 mscoco image captioning challenge,'' {\em IEEE
  transactions on pattern analysis and machine intelligence}, vol.~39, no.~4,
  pp.~652--663, 2017.

\bibitem{szegedy2016rethinking}
C.~Szegedy, V.~Vanhoucke, S.~Ioffe, J.~Shlens, and Z.~Wojna, ``Rethinking the
  inception architecture for computer vision,'' in {\em Proceedings of the IEEE
  Conference on Computer Vision and Pattern Recognition}, pp.~2818--2826, 2016.

\bibitem{Lin2014}
T.-Y. Lin, M.~Maire, S.~Belongie, J.~Hays, P.~Perona, D.~Ramanan,
  P.~Doll{\'a}r, and C.~L. Zitnick, ``Microsoft {COCO}: {C}ommon objects in
  context,'' in {\em European Conference on Computer Vision}, pp.~740--755,
  Springer, 2014.

\bibitem{repo:im2txt}
Google, ``Tensorflow model zoo: Im2txt.''

\bibitem{britz2017massive}
D.~Britz, A.~Goldie, M.-T. Luong, and Q.~Le, ``Massive exploration of neural
  machine translation architectures,'' in {\em Proceedings of the 2017
  Conference on Empirical Methods in Natural Language Processing},
  pp.~1442--1451, 2017.

\bibitem{Luong2015effective}
T.~Luong, H.~Pham, and C.~D. Manning, ``Effective approaches to attention-based
  neural machine translation,'' in {\em Proceedings of the 2015 Conference on
  Empirical Methods in Natural Language Processing}, pp.~1412--1421, 2015.

\bibitem{repo:nmt}
Google, ``Seq2seq: Neural machine translation.''

\bibitem{cublas}
NVIDIA, ``Cublas: Dense linear algebra on gpus.''

\bibitem{cutlass}
NVIDIA, ``Cutlass: Fast linear algebra in cuda c++.''

\end{thebibliography}

\end{document}